# SpaDeLeF: A Dataset for Hierarchical Classification of Lexical Functions for Collocations in Spanish

Yevhen Kostiuk[1], Grigori Sidorov[2], and Olga Kolesnikova[3]

Instituto Politécnico Nacional (IPN), Centro de Investigación en Computación (CIC),
Av. Juan de Dios Batiz, s/n, 07320, Mexico City, Mexico

[1]kosteugeneo@gmail.com; [2]sidorov@cic.ipn.mx; [3]kolesolga@gmail.com;

**Abstract.** In natural language processing (NLP), lexical function is a concept to unambiguously represent semantic and syntactic features of words and phrases in text first crafted in the Meaning-Text Theory. Hierarchical classification of lexical functions involves organizing these features into a tree-like hierarchy of categories or labels. This is a challenging task as it requires a good understanding of the context and the relationships among words and phrases in text. It also needs large amounts of labeled data to train language models effectively. In this paper, we present a dataset of most frequent Spanish verb-noun collocations and sentences where they occur, each collocation is assigned to one of 37 lexical functions defined as classes for a hierarchical classification task. Each class represents a relation between the noun and the verb in a collocation involving their semantic and syntactic features. We combine the classes in a tree-based structure, and introduce classification objectives for each level of the structure. The dataset was created by dependency tree parsing and matching of the phrases in Spanish news. We provide baselines and data splits for each objective.

## 1 Introduction

Hierarchical classification is a type of machine learning task where the goal is to classify instances into a tree-like hierarchy of categories or labels. In Natural Language Processing (NLP), hierarchical classification can be used for tasks such as text classification, where each text can be classified into multiple levels of categories or labels (Stein *et al.* 2019, Sajid *et al.* 2023). For example, a news article can be classified into such categories as sports, politics, and entertainment, and then further classified into such subcategories as football, international relations, and movies, respectively. In this way, hierarchical classification can help to organize and structure large amounts of text data.

Collocations are combinations of words frequently used in language; like other types of fixed multiword expressions, they create a challenge in the NLP field because of their complexional and idiomatic nature (Contreras Kallens & Christiansen 2022), even for large-scale language models (Wilkens *et al.* 2023). Being able to incorporate them in computer systems that deal with semantics should help the latter to avoid misunderstandings of provided texts and increase performance.

Up to date, a number of algorithms and approaches have been developed to address the issue of collocations (Espinosa-Anke *et al.* 2022, Deng & Liu 2022, Bisht *et al.* 2023, Simon 2023). Also, databases and machine-readable dictionaries of collocations manually annotated with grammatical and semantic information have been compiled, which can subsequently be used in a variety of applications (Chiarcos *et al.* 2022, Ottaiano & de Oliveira 2022, Reznowski 2023, Shabani & Dogolsara 2023), but such repositories are not sufficiently big to be used in robust systems, so semantic comprehension of collocations in language models remains a challenge. Therefore, strong algorithms capable of deep understanding of collocations as well as the meaning of free word combinations are required.

The issue of collocations becomes still more complex as they possess another feature beside idiomaticity: collocations are characterized by lexical diversity; it means that different words are used to lexicalize a single meaning. As an example, let us consider collocations *big need*, *breathtaking speed*, *deep love*, *fierce combat*, *infinite patience*. Although the first word in these collocations is different, their core meaning is the same and can be interpreted as 'big'. Moreover, the lexical, semantic and syntactic relations between the two words in all collocation are the same. Such relations are abstracted by the concept of lexical function, which in this example is Magn, from Latin *magnus*, big.

Therefore, the above given collocations can be formally represented as Magn(*need*) = *big*, Magn(*speed*) = *breathtaking*, Magn(*love*) = *deep*, Magn(*combat*) = *fierce*, Magn(*patience*) = *infinite*. Here the argument of Magn is the noun, the base in these collocations, and the value of Magn is the adjective, the collocator. Base and collocator are common terms to denote the elements of a collocation: the base is the head, used in its typical sense, and the collocator is semantically and syntactically dependent on the base, usually acquiring a sense different from its typical meaning. Let us take another example, *pay attention*: here *attention* is the base and keeps its typical meaning ('the act or state of applying the mind to something'[1]), and *pay* is the collocate and changes the meaning from 'to make due return to for services rendered or property delivered' to 'give'[2].

Magn is a lexical function, able, on the one hand, to identify similarity in diversity, and on the other hand, to return the correct word for each argument thus fixing its selectional preference. The latter will enrich language knowledge acquired by (large) language models to help them reduce the error rate they show now in tasks like machine translation (Borji 2023, Sholikhah & Indah 2021, Costa *et al.* 2015) and sentiment analysis (Bisht *et al.* 2023). Also, knowledge of collocation is indispensable in automatic text generation and in second language learning to produce naturally-sounded speech (Abdullayeva 2023, Kurniawan & Abdurrahim 2023). In such tasks as well as in many other areas of natural language processing, lexical functions will serve as a valuable tool to systematize and represent semantic and syntactic patterns of collocations.

More that 60 lexical functions have been defined formalize collocational knowledge (Mel'čuk 2015). In this work, we deal with lexical functions in verb-noun collocations such as *pay attention* exemplified previously. Section 2 specifies and gives definitions of these functions.

Verb-noun collocations must be distinguished from other verb-noun phrases which are free word combinations as they have different semantic properties. In free word combinations, the meaning of the phrase can be delivered from the conjoined meanings of its elements, however, it is not the case for collocations. For example, *pay the rent* is a free word combination, as its meaning can be derived from the words *pay* and *rent*. But in *pay attention* the meaning cannot be understood as a joint meaning of its elements, therefore, it is a collocation.

Nowadays, machine learning algorithms, especially artificial neural networks are the most utilized computational tools for text processing. In order to train them, it is necessary to have big datasets. Dataset compilation and annotation, which is normally done manually for a better quality, is a time-consuming and labor-intensive operation. Such resources are often lacking in many circumstances, particularly, for languages other than English.

Our goal was to contribute to the solution of the problem of lexical resources shortage by building a collection of Spanish verb-noun collocations annotated with lexical functions and used in their context extracted from a collection of 1,131 issues of the newspaper *Excelsior and* from the Spanish news dataset available on Kaggle[3] containing 58,424 news obtained from *La*

---

[1] https://www.merriam-webster.com/dictionary/attention
[2] https://www.merriam-webster.com/dictionary/pay
[3] https://www.kaggle.com/datasets/josemamuiz/noticias-laraznpblico/

*Razón*[4] (31,477 news) and *Público*[5] (26,948 news) Spanish newspapers.

This article is structured as follows: Section 2 describes lexical functions in our dataset, Section 3 explains the dataset construction, Section 4 speaks about our hierarchical classification of collocations, Section 5 presents the baseline results for the hierarchical classification, Section 6 discusses the results, and Section 7 concludes the article.

## 2 Lexical Functions for Verb-Noun Collocations in our Dataset

To compile our dataset, we used the list of Spanish verb-noun collocations annotated with lexical functions[6] (Gelbukh & Kolesnikova 2012). In this section we describe lexical functions in our dataset, and its compilation is discussed in Section 3.

The concept of lexical function (LF) was explained in the Introduction, here we provide more detail. First, the names of lexical functions are shortened versions of Latin words, selected with the same meaning as the respective LF, usually such names are self-explicative.

Table 1. Basic lexical functions in our dataset

| Lexical Function | Meaning | Examples |
|---|---|---|
| Anti | opposite, negation | *fail an examination*, *reject a piece of advice*, *turn down an application* |
| Caus | cause | *bring something under one's control*, *create a difficulty*, *hold an election* |
| Cont | continue | *maintain enthusiasm*, *hope burns* |
| Copul | linking word, copula | *be happy*, *have written* (Copul values underlined) |
| Fin | cease, finish | *lose patience*, *quench a desire* |
| Func | function, realize itself | *snow falls*, *the war is on* |
| Incep | begin, start | *acquire popularity*, *sink into despair*, *contract a disease* |
| Liqu | liquidate, abort | *withdraw support*, *divert attention* |
| Manif | manifest, show, exhibit | *amazement lurks* (*in his eyes*), *joy explodes* (*in her heart*), *scorn is dripping* (*from every word*) |
| Minus | decrease | *health fails*, *blow softens* |
| Oper | do, carry out, perform | *receive support*, *give an order* |
| Perf | perfect | *reach a grade*, *take measure*s |
| Perm | permit, allow | to give in to the desire |
| Plus | increase | |
| Real | fulfill the typical purpose of the event, expressed by the noun | *apply measure*, *fix a problem* |

Second, lexical functions can be simple and compound. A simple LF represents a single semantic unit and is denoted with an abbreviated Latin word reflecting the function's meaning. A compound LF includes more than one semantic unit. For example, Oper (Latin, *operor*, perform) and Incep (Latin, *incepere,* begin) are simple LFs meaning to perform and to begin, respectively. They are used to construct a compound LF IncepOper meaning to begin to perform (an action), e.g., as in *acquire a habit*, *run into trouble*.

Finally, LFs describe not only semantics in collocations, specifically for verb-noun collocations in our dataset, but also the syntactic relations among collocational elements using subscript numbers to identify semantic roles of the arguments in the verb's subcategorization frame. The number 1 denotes the agent, 2 is used for the recipient, 3 for the patient, and the order of the numbers explains the syntactic functions of the semantic roles. For example, Oper1 means to perform an action, the agent is the subject in sentences where Oper1 is used: *The professor applied the exam*. In Oper2, the patient of the action is the subject: *The student passed the exam*. Oper12 means that the subject in a sentence is the agent, and the recipient is the object: "*I feel enormous sympathy for people that live in poverty and fear.*"[7] If the number is zero, there is no agent neither recipient, the action realizes itself, e.g., Func0 in *snow falls*, see Table 1 for the definition of Func. Further details and discussion on the LF notation and meaning can be found in (Mel'čuk 2015).

---
[4] https://www.larazon.es/
[5] https://www.publico.es/
[6] http://www.gelbukh.com/lexical-functions/

Now we explain lexical functions found in our dataset using two tables. Table 1 lists simple LFs without their subcategorization notation to explain their core meaning and give examples. Table 2 uses the simple LFs in Table 1 to present the full notation of LFs in our dataset, also with their meaning and examples.

Table 2. Lexical functions in Spanish verb-noun collocations in our dataset, in their full notation. For each lexical function, we give its description, an example of a collocation from our dataset, and its English translation

| Lexical Function | Description | Examples | |
|---|---|---|---|
| | | Spanish | English translation |
| AntiReal3 | Failure to fulfill the typical purpose of the event (noun) with respect to the patient of the action (verb) | *violar el derecho* | violate the right |
| Caus1Func1 | Causation of the realization of the event (noun) by the agent | *sacar provecho* | take advantage |
| Caus1Oper1 | Causation of the event (noun) by the agent | *dar un resultado* | give a result |
| Caus2Func1 | Experiencing of the event (noun) caused by a non-agent of the situation | *dar miedo* | cause fear |
| CausFunc0 | Existence of an entity (noun) caused by an unidentified participant of the situation | *el plan se elabora* | the plan is developed |
| CausFunc1 | Existence of an entity (noun) caused by the agent | *ofrecer servicio* | provide a service |
| CausManifFunc0 | Existence and exhibition of an entity (noun) caused by an unidentified participant of the situation | *el concurso se anuncia* | the competition is advertised |
| CausMinusFunc0 | Decrease of the realization of an entity (noun) caused by an unidentified participant of the situation | *el riesgo se reduce* | the risk is reduced |
| CausMinusFunc1 | Decrease of the realization of an entity (noun) caused by the agent | *reducir el número* | reduce the number |
| CausPerfFunc0 | Existence and complete realization of an entity (noun) caused by an unidentified participant of the situation | *el derecho se garantiza* | the right is guaranteed |
| CausPlusFunc0 | Increasing realization of an entity (noun) caused by an unidentified participant of the situation | *el desarrollo se favorece* | the development is favored |
| CausPlusFunc1 | Increase of the realization of an entity (noun) caused by the agent | *promover el desarrollo* | promote the development |
| ContOper1 | Continuation of performing the event (noun) by the agent | *mantener la relación* | keep the relation |
| Copul | Linking verb | *ser parte* | be a part of |

---

[7] https://www.europarl.europa.eu/doceo/document/CRE-6-2006-04-06_EN.html?redirect/

| Lexical Function | Description | Examples | |
|---|---|---|---|
| | | Spanish | English translation |
| FinFunc0 | Termination of the realization of an event (noun) | *el plazo transcurre* | the time period elapsed |
| FinOper1 | Termination of the realization of an event (noun) by the agent | *perder control* | lose control |
| Func0 | Realization of an event (noun) | *tiempo pasó* | time passed |
| Func1 | Realization of an event (noun) by the agent | *(me) quedó duda* | a doubt remained |
| IncepFunc0 | Commencement of realization of an event (noun) | *la hora llega* | the hour comes |
| IncepOper1 | Commencement of realization of an event (noun) by the agent | *iniciar una sesión* | start a session |
| IncepReal1 | Commencement of realization of the typical purpose an event (noun) by the agent | *abordar un problema* | attack a problem |
| LiquFunc0 | Abortion of the realization of an event (noun) | *el problema se evita* | the problem is avoided |
| Manif | exhibition of an event (noun) | *mostrar interés* | show interest |
| ManifFunc0 | Existence and exhibition of an entity (noun) | *la pregunta se plantea* | the question is raised |
| MinusReal1 | Decrease of realization of the typical purpose an event (noun) by the agent | *gastar dinero* | spend money |
| Oper1 | Perform an event (noun) by the agent | *prestar atención* | pay attention |
| Oper2 | Experiencing an event (noun) by the recipient | *recibir atención* | receive attention |
| Oper3 | Experiencing an event (noun) by the patient | *contener información* | contain information |
| PerfFunc0 | Complete realization of an event (noun) | *el momento llega* | the moment comes |
| PerfOper1 | Perform an event (noun) to its full extent by the agent | *tomar precaución* | take precaution |
| PermOper1 | Allow to perform an event (noun) by the agent | *permitir acceso* | permit access |
| Real1 | Fulfillment of the typical purpose of the event (noun) with respect to the agent | *contestar una pregunta* | answer a question |
| Real2 | Fulfillment of the typical purpose of the event (noun) with respect to the recipient | *merecer atención* | deserve attention |
| Real3 | Fulfillment of the typical purpose of the event (noun) with respect to the patient | *reconocer el derecho* | recognize the right |

## 3 Dataset Construction

Our dataset can be accessed online[8] together with the code[9] for hierarchical classification described in Section 4. The dataset is structured as follows. First, as mentioned in Section 2, we used a list of collocations in Spanish manually gathered and described in (Gelbukh & Kolesnikova 2012) to build our dataset. This list contains 957 most frequent verb-noun collocations as well as free word combinations labeled as FWC found in the Spanish Web Corpus located in Sketch Engine[10] (Kilgarriff *et al*. 2014). FWCs were included in our dataset to train a language model to distinguish among collocations and free word combinations as a first step in the hierarchical classification of lexical functions (see Section 4). For every collocation, we retrieved its respective lexical function label.

---

[8] Under blind review.
[9] Under blind review.
[10] https://www.sketchengine.eu/

Second, for each collocation and FWC in the list mentioned above (Gelbukh & Kolesnikova 2012), we extracted all sentences with its occurrence by parsing (1) the text of 1,131 issues of the Mexican newspaper *Excelsior*[11] published within the period from April 01, 1996 to June 24, 1999 and (2) the text of Spanish news dataset available on Kaggle containing 58,424 news extracted from *La Razón*[12] (31,477) and *Público*[13] (26,948) Spanish newspapers. Specifically, the text was split into sentences, then for every sentence, the syntax tree was built using spaCy library for Python (Honnibal *et al.* 2020). A sentence was considered to have an occurrence of a collocation or an FWC if it complied with the following rules:

- Both the verb and the noun of a collocation or an FWC are present in the sentence;
- The noun is present in the verb list of the children of the syntax tree or the verb is present in the noun list of the children of the syntax tree.

Such rules secured that if the noun and the verb of a collocation or an FWC are present in the sentence, then they indeed form a phrase, not just "stand nearby" without any syntactic relation. As Spanish is a language with reach morphology and a flexible word order, it is difficult to determine syntactic relations without dependency parsing.

Third, we assigned a class label to each parsed sentence (the respective lexical function label or the FWC label depending on the phrase found in it), as well as marked the collocation or the FWC within the sentence.

Table 3 presents the dataset statistics: the number of sentences (# sentences), tokens (#tokens), unique tokens (# unique tokens) and lemmas (# lemmas) for each lexical function and FWC and the overall number of unique collocations and FWC (# phrases) found in the sentences. It can be noted in the table that the dataset is not balanced both in terms of the number of parsed sentences as well as the number of collocations per lexical function.

Table 3. Dataset statistics

| Lexical function | # phrases | # sentences | # tokens | # unique tokens | # lemmas | Average sentence length |
|---|---|---|---|---|---|---|
| AntiReal3 | 1 | 244 | 13,367 | 3,406 | 2,690 | 54.783 |
| Caus1Func1 | 3 | 1,204 | 56,221 | 9,153 | 6,816 | 46.695 |
| Caus1Oper1 | 2 | 2,448 | 116,722 | 14,422 | 10,648 | 47.681 |
| Caus2Func1 | 16 | 5,997 | 307,657 | 27,070 | 19,668 | 51.302 |
| CausFunc0 | 112 | 48,317 | 2,486,753 | 78,664 | 59,681 | 51.467 |
| CausFunc1 | 90 | 52,860 | 2,869,884 | 79,262 | 59,323 | 54.292 |
| CausManifFunc0 | 2 | 248 | 13,952 | 2,919 | 2,277 | 56.258 |
| CausMinusFunc0 | 3 | 1,160 | 46,975 | 7,335 | 5,492 | 40.496 |
| CausMinusFunc1 | 1 | 544 | 27,327 | 5,355 | 4,136 | 50.233 |
| CausPerfFunc0 | 1 | 696 | 39,250 | 5,778 | 4,267 | 56.393 |
| CausPlusFunc0 | 7 | 2,401 | 123,515 | 13,491 | 9,721 | 51.443 |
| CausPlusFunc1 | 5 | 2,735 | 125,156 | 13,634 | 10,092 | 45.761 |
| ContOper1 | 16 | 10,110 | 557,304 | 26,212 | 18,852 | 55.124 |
| Copul | 9 | 946 | 41,881 | 7,655 | 5,817 | 44.272 |
| FWC | 196 | 96,213 | 4,576,509 | 104,236 | 79,807 | 47.566 |
| FinFunc0 | 1 | 64 | 3,833 | 1,095 | 928 | 59.891 |

---

[11] https://www.publico.es/

[12] https://www.larazon.es/

[13] https://www.publico.es/

| Lexical function | # phrases | # sentences | # tokens | # unique tokens | # lemmas | Average sentence length |
|---|---|---|---|---|---|---|
| FinOper1 | 6 | 1,898 | 91,737 | 12,847 | 9,737 | 48.334 |
| Func0 | 25 | 50,041 | 2,349,590 | 81,042 | 61,239 | 46.953 |
| Func1 | 4 | 2,306 | 109,083 | 13,606 | 9,953 | 47.304 |
| IncepFunc0 | 3 | 2,022 | 98,457 | 13,530 | 10,118 | 48.693 |
| IncepOper1 | 25 | 16,161 | 795,702 | 41,029 | 29,628 | 49.236 |
| IncepReal1 | 2 | 509 | 25,047 | 4,605 | 3,492 | 49.208 |
| LiquFunc0 | 2 | 426 | 22,232 | 4,837 | 3,784 | 52.188 |
| Manif | 13 | 3,749 | 292,872 | 23,681 | 17,380 | 53.182 |
| ManifFunc0 | 1 | 111 | 6,435 | 1,895 | 1,521 | 57.973 |
| MinusReal1 | 1 | 310 | 15,116 | 3,542 | 2,768 | 48.761 |
| Oper1 | 279 | 212,599 | 11,040,533 | 150,675 | 120,017 | 51.932 |
| Oper2 | 30 | 8,761 | 434,646 | 32,857 | 24,118 | 49.611 |
| Oper3 | 1 | 182 | 11,937 | 2,808 | 2,266 | 65.588 |
| PerfFunc0 | 1 | 2,939 | 121,386 | 10,893 | 7,930 | 41.302 |
| PerfOper1 | 4 | 3,272 | 162,319 | 16,187 | 11,561 | 49.608 |
| PermOper1 | 3 | 670 | 36,312 | 6,876 | 5,271 | 54.197 |
| Real1 | 61 | 28,240 | 1,420,143 | 56,731 | 41,519 | 50.288 |
| Real2 | 3 | 2,942 | 137,612 | 14,188 | 10,343 | 46.775 |
| Real3 | 1 | 1,398 | 95,973 | 5,737 | 4,327 | 68.650 |

## 4 Hierarchical Classification

We propose several classification tasks, which form a tree structure, see Figure 1. Level 1 of the classification is to distinguish among lexical functions (LF) and free word combinations (FWC). At Level 2, we classified collocations into the following ten categories: Caus, ContOper, Copul, Func, Fin, Incep, Manif, Oper, Perf, Real. At Level 3, five categories of Level 2 were further classified as follows since they include more specific lexical functions:

- Caus was classified into four classes: (1) Caus1Func1, CausPlusFunc1, CausManifFunc0, CausPlusFunc0, CausPerfFunc0, CausMinusFunc0, and CausMinusFunc1, (2) CausFunc0, (3) CausFunc1, (4) Caus2Func1.
- Func was classified into two classes: (1) Func0, (2) Func1.
- Incep was classified into two classes: (1) IncepReal1 and IncepFunc0, (2) IncepOper1.
- Oper was classified into two classes: (1) Oper2 and Oper3, (2) Oper1.
- Real was classified also into two classes: (1) Real2 and Real 3, (2) Real1.

As it can be seen in the above classification at Level 3, some classes include more than one lexical function. This is due to a small number of collocations and sentences in our dataset for such lexical functions as well as to their similarity.

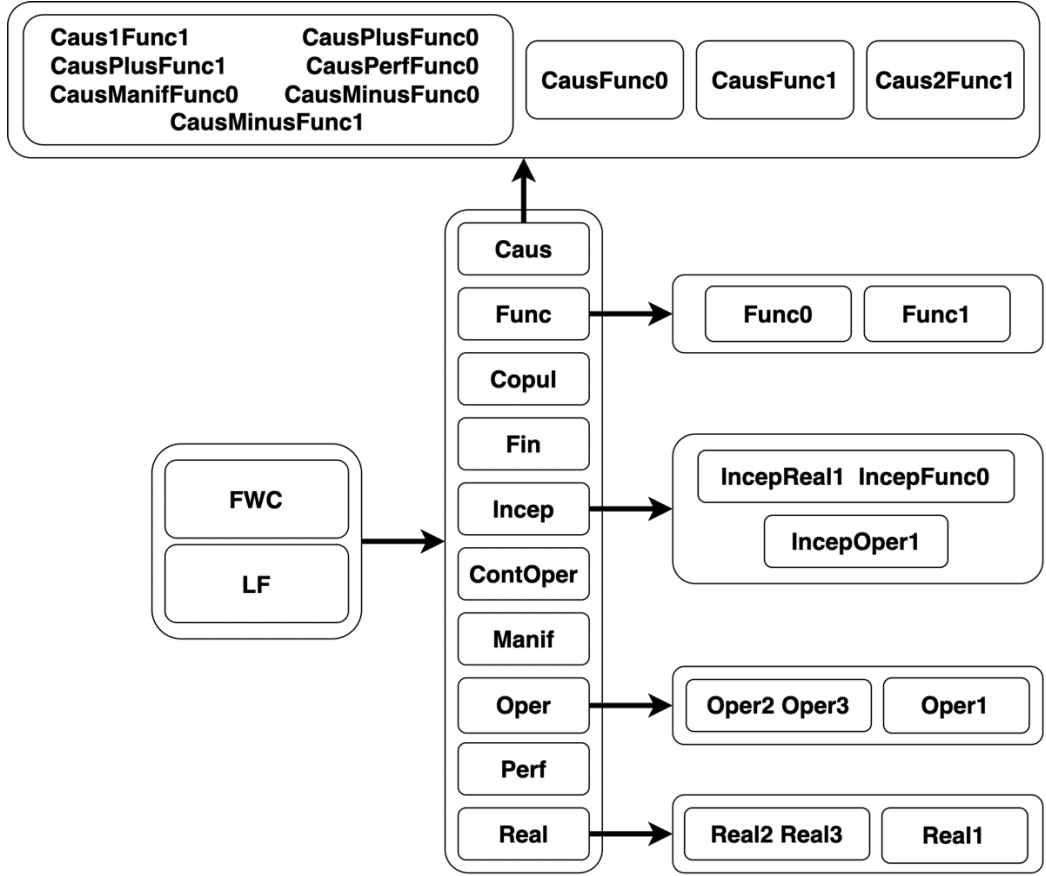

Figure 1. Hierarchical classification of verb-noun phrases

As a validation procedure, we selected a *k*-fold validation technique, with two or three folds depending on the size of the dataset for a given lexical function at a particular level of classification. The folds split was made based on the number of phrases (collocations or FWCs), not on the number of sentences for such phrases. It means that if a phrase was considered to be a part of the training folds, then all the sentences with such phrase were included in the training folds. The same is true for the test fold. Due to a big difference in the number of collocations per lexical function as well as the number of sentences per collocation or free word combination, we report the dataset statistics and the classification results for each fold in Section 5. In order to confront over-fitting or "memorizing" phrases, we masked the collocation or FWC in text for the algorithm chosen for classification at each level to focus on the context and its semantics.

As our baselines at Levels 1 and 2 of the hierarchical classification, we used BETO, a transformer model trained on Spanish text (Cañete *et al*. 2023), specifically, the BETO's version for sentence similarity available on the Hugging Face ecosystem[14]. Our choice of this model is based on its excellent performance on NLP tasks for Spanish (Inácio & Oliveira 2023, López-Ávila *et al*. 2023, Meza Lovon 2023, Rubio *et al*. 2023). To classify lexical functions at Level 3, we selected two classical machine learning algorithms which proved their high performance on the classification task: Naïve Bayes (NB) (Dawar & Kumar 2023, Shabani *et al*. 2023) and Support Vector Machine (SVM) (Gasparetto *et al*. 2022, Hassan *et al*. 2022).

In the stage of text preprocessing and feature extraction, we performed two steps: first, tokenization in words and second, lemmatization using spaCy Python package

---

[14] https://huggingface.co/hiiamsid/sentence_similarity_spanish_es

([Honnibal *et al.* 2020](#)). Classification of phrases into collocations and FWC and further into lexical function was based on the context of phrases, i.e., no lexical knowledge from dictionaries or other sources was used. To extract the context of each phrase, we defined a specific window size. For example, consider the following sentence:

*Hace unas semanas, fue a la LIX convención de la banca, en Cancún.*
A few weeks ago, (he/she) went to the LIX banking convention in Cancun.

After tokenization and lemmatization, the sentence looks like this:

*hacer*, *un*, *semana*, *ser*, *a*, *el*, *lix*, *convención*, *de*, *el*, *banca*, *en*, *cancún*.

At Levels 1 and 2 we used the whole sentence as input to BETO, and at Level 3, as initial input features to NB and SVM, the right and left tokens in the context of the collocation are selected. In this example, the collocation is *hacer semana* (a week ago)[15]. For window size 2, the input to the algorithm will be 2 tokens of the right context: *ser*, *a*. As *hacer* is the first token in the sentence, the right window is empty. In our experiment we checked several window sizes[16] in the range from 4 to 20, then choosing the best performing window size to report our results in Section 5. After obtaining the initial input context tokens, we used TF-IDF vectorization to generate numeric input features to feed into the selected classification algorithms. So, for example, the combination *hacer semana* is classified progressively depending on the level of the hierarchy as LF (Level 1), Func (Level 2), and Func0 (Level 3).

To evaluate the performance of the selected algorithms, we applied precision, recall, and F1-score. In Section 5 we provide results per fold and their weighted average over all folds.

## 5 Baseline Results

In this section, we present the results of our selected baseline methods: BETO (Levels 1 and 2), Naïve Bayes and Support Vector Machine (Level 3). As we mentioned in Section 4, we experimented with several window sizes at Level 3 and the best results for all LFs were obtained for the window of 4, except for Real functions, where the algorithms performed best on window size 20. Also, in Section 4 we explained the reason for which we present the data statistics and the classification results for each fold: this is due to a big difference in the number of collocations per lexical function and the number of sentences per FWC or collocation, observe this diversity in Table 3.

### 5.1 Classification at Level 1

At Level 1 we experimented with BETO to classify all phrases into two classes: (1) collocations of any lexical function and (2) free word combinations (FWC). Section 4 introduces BETO briefly, and here we give more detail on the model's architecture: we used randomly selected batches of 128 samples, embedding size of 768, hidden layer size of 256, and trained the model on 3,500 iterations.

Table 4 shows the dataset statistics per fold, and Table 5 includes the results for each fold and class. As input to the BETO model, we used the whole sentence masking the verb of a collocation or free word combination in the sentence and training the model for binary

---

[15] The English equivalent of the Spanish phrase *hace una semana* is *a week ago*. Syntactically, the English translation is different from Spanish: *ago* in *a week ago* is an adverb, but *hace* in *hace una semana* is a verb, its lemma is *hacer*, therefore, the preprocessed form of *hace una semana* is *hacer semana*.

[16] Window size *n* means *n* context words to the left of a given word and *n* context words to the right.

classification.

Table 4. Dataset statistics for classifying verb-noun phrases into two classes: collocation of any lexical function (LF) vs free word combination (FWC)

| Class label | Fold | Train | | Test | |
|---|---|---|---|---|---|
| | | # phrases | # sentences | # phrases | # sentences |
| LF | 1 | 492 | 297,038 | 246 | 173,230 |
| FWC | | 134 | 51,782 | 68 | 44,431 |
| LF | 2 | 493 | 310,144 | 245 | 160,124 |
| FWC | | 134 | 69,945 | 68 | 26,268 |
| LF | 3 | 491 | 333,354 | 247 | 136,914 |
| FWC | | 136 | 70,699 | 66 | 25,514 |

Table 5. Detailed results for classifying verb-noun phrases into two classes: collocation of any lexical function (LF) vs free word combination (FWC) using BETO

| Class label | Fold | Precision | Recall | F1-score |
|---|---|---|---|---|
| LF | 1 | 0.811 | 0.575 | 0.673 |
| FWC | | 0.224 | 0.478 | 0.305 |
| macro average | | 0.517 | 0.526 | 0.489 |
| weighted average | | 0.691 | 0.555 | 0.598 |
| LF | 2 | 0.889 | 0.715 | 0.793 |
| FWC | | 0.208 | 0.456 | 0.286 |
| macro average | | 0.549 | 0.586 | 0.539 |
| weighted average | | 0.793 | 0.678 | 0.721 |
| LF | 3 | 0.870 | 0.697 | 0.774 |
| FWC | | 0.213 | 0.440 | 0.287 |
| macro average | | 0.541 | 0.568 | 0.530 |
| weighted average | | 0.767 | 0.657 | 0.697 |
| macro average of folds | | 0.535 | 0.560 | 0.519 |
| weighted average of folds | | 0.750 | 0.630 | 0.672 |

It can be noted that lexical functions are classified with high precision and recall on fold 2: 0.889 and 0.715, respectively. This shows that collocations are discriminated well from free word combinations, however, if the purpose is to detect free word combinations as opposite to collocations, the performance drops significantly with a precision as low as 0.208 and a recall of 0.456 on the same fold 2. The best F1-score is 0.793 for LF and the best weighted average F1-score for both classes is 0.721 on fold 2.

**5.2 Classification at Level 2**

At Level 2, we deal with collocations only. The purpose here is to distinguish among ten classes represented by the following lexical function types: Caus, ContOper1, Copul, Fin, Func, Incep, Manif, Oper, Perf, and Real. Two of these classes include specific lexical functions (ContOper1, Copul), the rest result from grouping various specific lexical functions with the same core meaning: e.g., CausFunc0, CausFunc1, Caus1Func1, Caus2Func1, CausPlusFunc0, CausPlusFunc1, CausPerfFunc0, CausMinusFunc0, CausMinus Func1, and CausManifFunc0 are grouped under the umbrella type Caus; Oper1, Oper2, and Oper3 are of type Oper.

Table 6 shows the dataset statistics for the classification, and Table 7 details the results for each fold, class, and algorithm. At this level we used BETO with the same configuration as that of Level 1, see Section 5.1.

Table 6. Dataset statistics for classifying collocations in ten classes, each class is a specific lexical function type

| Class label | Fold | Train | | Test | |
|---|---|---|---|---|---|
| | | # collocations | # sentences | # collocations | # sentences |
| Caus | 1 | 164 | 74,938 | 78 | 43,672 |
| ContOper1 | | 10 | 8,170 | 6 | 1,940 |
| Copul | | 8 | 647 | 1 | 299 |
| Fin | | 3 | 1,477 | 4 | 485 |
| Func | | 20 | 40,765 | 9 | 11,582 |
| Incep | | 22 | 12,977 | 8 | 5,715 |
| Manif | | 9 | 4,116 | 5 | 5,618 |
| Oper | | 211 | 147,812 | 101 | 73,730 |
| Perf | | 1 | 3,266 | 4 | 2,945 |
| Real | | 38 | 20,217 | 27 | 12,363 |
| Caus | 2 | 165 | 84,405 | 77 | 34,205 |
| ContOper1 | | 12 | 5,688 | 4 | 4,422 |
| Copul | | 4 | 688 | 5 | 258 |
| Fin | | 5 | 1,060 | 2 | 902 |
| Func | | 21 | 34,193 | 8 | 18,154 |
| Incep | | 23 | 14,751 | 7 | 3,941 |
| Manif | | 9 | 3,060 | 5 | 2,558 |
| Oper | | 196 | 139,169 | 116 | 82,373 |
| Perf | | 3 | 3,130 | 2 | 3,081 |
| Real | | 44 | 25,626 | 21 | 6,954 |
| Caus | 3 | 155 | 77,877 | 87 | 40,733 |
| ContOper1 | | 10 | 6,359 | 6 | 3,748 |
| Copul | | 6 | 557 | 3 | 389 |
| Fin | | 4 | 1,387 | 3 | 575 |
| Func | | 17 | 45,775 | 12 | 6,572 |
| Incep | | 15 | 9,656 | 15 | 9,036 |
| Manif | | 10 | 4,562 | 4 | 1,056 |
| Oper | | 217 | 156,103 | 95 | 65,439 |
| Perf | | 4 | 6,026 | 1 | 185 |
| Real | | 48 | 19,317 | 17 | 13,263 |

Table 7. Results for classifying collocations in ten classes using BETO, best results in bold

| Class label | Fold | Precision | Recall | F1-score |
|---|---|---|---|---|
| Caus | 1 | 0.318 | 0.161 | 0.214 |
| ContOper1 | | 0.009 | 0.031 | 0.014 |
| Copul | | 0.007 | 0.351 | 0.014 |
| Fin | | 0.008 | 0.155 | 0.014 |
| Func | | 0.283 | 0.343 | 0.310 |
| Incep | | 0.088 | 0.103 | 0.095 |
| Manif | | 0.336 | 0.407 | 0.062 |
| Oper | | 0.513 | 0.161 | 0.246 |
| Perf | | 0.038 | 0.178 | 0.063 |
| Real | | 0.170 | 0.341 | 0.227 |
| macro average | | 0.147 | 0.223 | 0.126 |
| weighted average | | 0.375 | 0.189 | 0.225 |
| Caus | 2 | 0.218 | 0.053 | 0.085 |
| ContOper1 | | 0.115 | 0.386 | 0.178 |
| Copul | | 0.004 | 0.252 | 0.009 |
| Fin | | 0.010 | 0.165 | 0.019 |
| Func | | 0.539 | 0.524 | 0.532 |
| Incep | | 0.034 | 0.166 | 0.067 |

| Class label | Fold | Precision | Recall | F1-score |
|---|---|---|---|---|
| Manif | | 0.068 | 0.365 | 0.115 |
| Oper | | 0.583 | 0.143 | 0.229 |
| Perf | | 0.020 | 0.506 | 0.028 |
| Real | | 0.085 | 0.291 | 0.131 |
| macro average | | 0.168 | 0.240 | 0.138 |
| weighted average | | 0.435 | 0.215 | 0.244 |
| Caus | 3 | 0.258 | 0.049 | 0.082 |
| ContOper1 | | 0.037 | 0.045 | 0.040 |
| Copul | | 0.010 | 0.026 | 0.020 |
| Fin | | 0.013 | 0.254 | 0.025 |
| Func | | 0.284 | 0.286 | 0.285 |
| Incep | | 0.126 | 0.388 | 0.190 |
| Manif | | 0.029 | 0.414 | 0.054 |
| Oper | | 0.485 | 0.199 | 0.283 |
| Perf | | 0.001 | 0.038 | 0.002 |
| Real | | 0.171 | 0.307 | 0.220 |
| macro average | | 0.141 | 0.224 | 0.120 |
| weighted average | | 0.338 | 0.180 | 0.203 |
| macro average of folds | | 0.152 | 0.229 | 0.128 |
| weighted average of folds | | 0.383 | 0.195 | 0.224 |

It can be seen in Table 7 that the results are very modest, the highest result is for Func in fold 2: 0.539, 0.524, 0.532 for precision, recall and F1-score, respectively. The highest weighted average F1-score is 0.244 in fold 1, which demonstrates a poor performance of BETO at this Level compared with Level 1 where BETO showed much higher ability to distinguish verb-noun phrases as LF or FWC with weighted average F1-score of 0.721 on fold 2.

Low results on classification leave room for future research to study BETO's operation on distinct phrases and sentences types for better understanding the advantages and limitations of the model. Another line of research will be testing other machine learning methods and language models on our dataset at this level of hierarchical classification or develop another classification paradigm.

### 5.3 Classification at Level 3

At Level 3, we use Naïve Bayes (NB) and Support Vector Machine (SVM) to further classify collocations in five out of ten classes distinguished at Level 2. We discussed our choice of NB and SVM for this level in Section 4. The five classes we work here are Caus, Func, Incep, Oper, and Real, each one includes specific lexical functions. The details on data and results for each class are presented in their respective subsections.

#### 5.3.1 Caus

Under the label Caus, we grouped LFs with the causation core meaning and distributed them among four classes. In the first class, we grouped seven LFs with causation semantics due to a small number of collocations in each LF, these functions are Caus1Func1, CausPlusFunc0, CausPlusFunc1, CausMinusFunc0, CausMinusFunc1, CausPerfFunc0, and CausManifFunc0. Even though such grouping does not result in a big number of collocations (altogether there are 45 samples), the total number of sentences for these seven LFs is substantial, see Table 3, so it made sense to include them in the classification scheme. Besides, their semantics is highly similar, basically, they differ with respect to their verb subcategorization frames.

The resting three classes are CausFunc0, CausFunc1, and Caus2Func1. The meaning of collocations in these classes differs from the meaning of the first class with LFs listed in the previous paragraph, but differences among the three classes are subtle. In spite of that, the number of instances in each class is sufficient for an attempt to distinguish among them, so we used them in our experiments. Table 8 shows the dataset statistics for the classification, Table 9 presents averaged results for all four classes, and Table 10 details the results for each fold, class, and machine learning technique.

Table 8. Dataset statistics for classifying Caus collocations into four classes

| Class label | Fold | Train | | Test | |
|---|---|---|---|---|---|
| | | # collocations | # sentences | # collocations | # sentences |
| CausFunc0 | 1 | 75 | 34,554 | 37 | 10,763 |
| CausFunc1 | | 58 | 28,907 | 32 | 23,953 |
| Caus2Func1 | | 11 | 4,787 | 5 | 1,210 |
| Caus1Func1 & other 6 Caus LFs | | 16 | 6,545 | 6 | 2,007 |
| CausFunc0 | 2 | 67 | 30,422 | 45 | 17,895 |
| CausFunc1 | | 67 | 44,781 | 23 | 8,079 |
| Caus2Func1 | | 11 | 3,883 | 5 | 2,114 |
| Caus1Func1 & other 6 Caus LFs | | 16 | 6,273 | 7 | 2,715 |
| CausFunc0 | 3 | 82 | 28,658 | 30 | 19,659 |
| CausFunc1 | | 55 | 32,032 | 35 | 20,828 |
| Caus2Func1 | | 10 | 5,093 | 6 | 904 |
| Caus1Func1 & other 6 Caus LFs | | 13 | 4,772 | 9 | 4,266 |

Table 9. Overall averaged results for classifying Caus collocations into four classes using Naïve Bayes (NB) and Support Vector Machine (SVM), best results in bold

| Model | Average weighted precision | Average weighted recall | Average weighted F1-score |
|---|---|---|---|
| NB | 0.432 | **0.331** | **0.356** |
| SVM | **0.456** | 0.269 | 0.306 |

Table 10. Results for classifying Caus collocations using Support Vector Machine (SVM) and Random Forest Classifier (RFC), best results in bold

| Class label | Fold | NB | | | SVM | | |
|---|---|---|---|---|---|---|---|
| | | Precision | Recall | F1-score | Precision | Recall | F1-score |
| CausFunc0 | 1 | 0.236 | 0.194 | 0.213 | 0.308 | 0.273 | 0.289 |
| CausFunc1 | | 0.630 | 0.474 | 0.541 | 0.611 | 0.184 | 0.283 |
| Caus2Func1 | | 0.038 | 0.175 | 0.063 | 0.037 | 0.407 | 0.067 |
| Caus1Func1 & other 6 Caus LFs | | 0.078 | 0.214 | 0.115 | 0.113 | 0.441 | 0.180 |
| macro average | | 0.246 | 0.265 | 0.233 | 0.267 | 0.326 | 0.205 |
| weighted average | | 0.470 | 0.372 | 0.410 | 0.480 | 0.230 | 0.272 |
| CausFunc0 | 2 | 0.599 | 0.315 | 0.412 | 0.607 | 0.308 | 0.408 |
| CausFunc1 | | 0.262 | 0.156 | 0.195 | 0.254 | 0.282 | 0.267 |
| Caus2Func1 | | 0.165 | 0.548 | 0.254 | 0.174 | 0.350 | 0.232 |
| Caus1Func1 & other 6 Caus LFs | | 0.096 | 0.194 | 0.128 | 0.148 | 0.367 | 0.210 |
| macro average | | 0.281 | 0.303 | 0.248 | 0.296 | 0.327 | 0.280 |
| weighted average | | 0.422 | 0.293 | 0.316 | 0.430 | 0.311 | 0.337 |
| CausFunc0 | 3 | 0.433 | 0.475 | 0.453 | 0.454 | 0.219 | 0.295 |
| CausFunc1 | | 0.448 | 0.211 | 0.287 | 0.536 | 0.253 | 0.343 |
| Caus2Func1 | | 0.019 | 0.125 | 0.033 | 0.033 | 0.532 | 0.062 |
| Caus1Func1 & other 6 Caus LFs | | 0.126 | 0.245 | 0.166 | 0.181 | 0.500 | 0.265 |
| macro average | | 0.257 | 0.264 | 0.235 | 0.301 | 0.376 | 0.242 |
| weighted average | | 0.403 | 0.327 | 0.342 | 0.457 | 0.267 | 0.310 |

Table 9 shows averaged results on classification in all classes and it can be noted that there is no single algorithm to classify Caus functions best. Although SVM gives best precision of 0.456, its recall of 0.269 is lower than the recall shown by NB of 0.33. The best weighted average F1-score of 0.356 is demonstrated by NB. Overall, the results are not high which leaves room for further research.

Observing the results per class in Table 10, we see that NB was able to distinguish CausFunc1 with a high precision of 0.630, its recall of 0.474 is lower, resulting in F1-score of 0.541 on fold 1, the best over all classes and folds. The best weighted average F1-score of 0.410 for NB is on fold 1. SVM showed best performance on CausFunc0 on fold 2 with 0.607, 0.308, 0,408 for precision, recall, and F1-score, respectively. The best weighted average F1-score for SVM is on fold 2.

### 5.3.2 Func

Func category includes two classes: Func0 and Func1. Table 11 shows the dataset statistics for the classification, Table 12 includes overall averaged classification results for both classes, and Table 13 details the results for each fold, class, and algorithm.

Table 11. Dataset statistics for classifying Func collocations in two classes

| Class label | Fold | Train | | Test | |
|---|---|---|---|---|---|
| | | # collocations | # sentences | # collocations | # sentences |
| Func0 | 1 | 13 | 32,842 | 12 | 17,199 |
| Func1 | | 3 | 1,374 | 1 | 932 |
| Func0 | 2 | 13 | 32,842 | 12 | 17,199 |
| Func1 | | 3 | 1,374 | 1 | 932 |

Table 12. Overall averaged results for classifying Func collocations in two classes using Naïve Bayes (NB) and Support Vector Machine (SVM), best results in bold

| Model | Average weighted precision | Average weighted recall | Average weighted F1-score |
|---|---|---|---|
| NB | **0.930** | 0.407 | 0.536 |
| SVM | 0.918 | **0.643** | **0.747** |

Table 13. Results for classifying Func collocations in two classes using Naïve Bayes (NB) and Support Vector Machine (SVM), best results in bold

| Class label | Fold | NB | | | SVM | | |
|---|---|---|---|---|---|---|---|
| | | Precision | Recall | F1-score | Precision | Recall | F1-score |
| Func0 | 1 | 0.969 | 0.419 | 0.585 | 0.965 | 0.645 | 0.773 |
| Func1 | | 0.047 | 0.680 | 0.087 | 0.050 | 0.447 | 0.090 |
| macro average | | 0.508 | 0.549 | 0.340 | 0.508 | 0.546 | 0.432 |
| weighted average | | 0.932 | 0.429 | 0.565 | 0.929 | 0.637 | 0.746 |
| Func0 | 2 | 0.974 | 0.361 | 0.527 | 0.952 | 0.663 | 0.782 |
| Func1 | | 0.065 | 0.824 | 0.121 | 0.058 | 0.385 | 0.101 |
| macro average | | 0.520 | 0.593 | 0.324 | 0.505 | 0.524 | 0.442 |
| weighted average | | 0.928 | 0.385 | 0.506 | 0.906 | 0.649 | 0.747 |

The average weighted results on Func classification are much higher that on the rest of classes at Level 3. Table 12 shows a precision of 0.930 for NB, however, its recall of 0.407 is much lower resulting in an F1-score of 0.536. In contrast, SVM give a higher recall of 0.643 but a lower precision of 0.918, and its F1-score is 0.747 which is higher than F1-score of 0.536 for NB.

Concerning results per class and fold, the best results for NB is on Func0 with 0.969, 0.419, 0.585 for precision, recall, and F1-score, respectively. It is interesting that SVM also detected Func0 better than Func1, but on fold 2 with precision, recall, and F1-score of 0.952, 0.663, and 0.782, respectively.

### 5.3.3 Incep

The Incep category includes two classes: the first class contains collocations of IncepReal1 and IncepFunc0 put together due to their similarity and little data: 2 collocations in 509 sentences for IncepReal1 and 3 collocations in 2,022 sentences for IncepFunc0, see Table 3. The second class is IncepOper1. Table 14 shows the dataset statistics for the classification, Table 15 includes overall averaged classification results for both classes, and Table 16 details the results for each fold, class, and algorithm.

Table 14. Dataset statistics for classifying Incep collocations in two classes

| Class label | Fold | Train | | Test | |
|---|---|---|---|---|---|
| | | # collocations | # sentences | # collocations | # sentences |
| IncepReal1 &IncepFunc0 | 1 | 4 | 1,973 | 1 | 558 |
| IncepOper1 | | 14 | 10,448 | 11 | 5,713 |
| IncepReal1 &IncepFunc0 | 2 | 4 | 1,973 | 1 | 558 |
| IncepOper1 | | 14 | 10,448 | 11 | 5,713 |

Table 15. Overall averaged results for classifying Incep collocations in two classes using Naïve Bayes (NB) and Support Vector Machine (SVM), best results in bold

| Model | Average weighted precision | Average weighted recall | Average weighted F1-score |
|---|---|---|---|
| NB | **0.781** | **0.603** | **0.667** |
| SVM | 0.776 | 0.508 | 0.566 |

Table 16. Results for classifying Incep collocations into two classes using Naïve Bayes (NB) and Support Vector Machine (SVM), best results in bold

| Class label | Fold | Naïve Bayes | | | SVM | | |
|---|---|---|---|---|---|---|---|
| | | Precision | Recall | F1-score | Precision | Recall | F1-score |
| IncepReal1 &IncepFunc0 | 1 | 0.302 | 0.529 | 0.384 | 0.296 | 0.269 | 0.282 |
| IncepOper1 | | 0.780 | 0.577 | 0.663 | 0.755 | 0.779 | 0.767 |
| macro average | | 0.541 | 0.553 | 0.524 | 0.526 | 0.524 | 0.524 |
| weighted average | | 0.657 | 0.565 | 0.592 | 0.637 | 0.648 | 0.642 |
| IncepReal1 &IncepFunc0 | 2 | 0.050 | 0.341 | 0.088 | 0.060 | 0.724 | 0.104 |
| IncepOper1 | | 0.949 | 0.657 | 0.777 | 0.959 | 0.348 | 0.510 |
| macro average | | 0.500 | 0.499 | 0.432 | 0.508 | 0.536 | 0.307 |
| weighted average | | 0.904 | 0.641 | 0.742 | 0.914 | 0.367 | 0.490 |

Table 15 presents the results averaged over classes and folds: the best overall average weighted F1-score of 0.667 was shown by NB, its precision is 0.781, and its recall is 0.603. It cannot be said that SVM's performance is low, it is not as high as that of NB, still it showed a precision of 0.776, a little lower than that of NB. Concerning classification per class and fold, NB detected IncepOper1 quite successfully, with a precision of 0.949, a recall of 0.657, and an F1-score of 0.777. SVM showed competing results for the same lexical function but in fold1, with a precision of 0.755, a recall of 0.779, and an F1-score of 0.767.

### 5.3.4 Oper

The Oper category includes two classes: the first class contains collocations of Oper2 and Oper3 united in one class because of their similarity and a small number of collocations: although Oper 2 has 30 collocations in 8,761 sentences, Oper 3 has only 1 collocation in 182 sentences, see Table 3. The second class is Oper1. Table 17 shows the dataset statistics for the classification, Table 18 includes overall averaged classification results for both classes, and Table 19 details the results for each fold, class, and algorithm.

Table 17. Dataset statistics for classification of Oper collocations into two classes

| Class label | Fold | Train | | Test | |
| --- | --- | --- | --- | --- | --- |
| | | # collocations | # sentences | # collocations | # sentences |
| Oper2 & Oper3 | 1 | 18 | 5,734 | 13 | 3,209 |
| Oper1 | | 190 | 151,325 | 91 | 61,274 |
| Oper2 & Oper3 | 2 | 20 | 4,906 | 11 | 4,037 |
| Oper1 | | 188 | 123,725 | 93 | 88,874 |
| Oper2 & Oper3 | 3 | 24 | 7,246 | 7 | 1,697 |
| Oper1 | | 184 | 150,148 | 97 | 62,451 |

Table 18. Overall averaged results for classifying Oper collocations in two classes using Naïve Bayes (NB) and Support Vector Machine (SVM), best results in bold

| Model | Average weighted precision | Average weighted recall | Average weighted F1-score |
| --- | --- | --- | --- |
| NB | 0.929 | **0.427** | **0.564** |
| SVM | **0.931** | 0.274 | 0.387 |

Table 19. Results for classification of Oper collocations into two classes using Naïve Bayes (NB) and Support Vector Machine (SVM), best results in bold

| Class label | Fold | Naïve Bayes | | | SVM | | |
| --- | --- | --- | --- | --- | --- | --- | --- |
| | | Precision | Recall | F1-score | Precision | Recall | F1-score |
| Oper2 & Oper3 | 1 | 0.055 | 0.638 | 0.101 | 0.055 | 0.810 | 0.104 |
| Oper1 | | 0.957 | 0.426 | 0.590 | 0.965 | 0.277 | 0.431 |
| macro average | | 0.506 | 0.532 | 0.346 | 0.510 | 0.544 | 0.267 |
| weighted average | | 0.913 | 0.437 | 0.566 | 0.920 | 0.304 | 0.414 |
| Oper2 & Oper3 | 2 | 0.048 | 0.641 | 0.090 | 0.045 | 0.784 | 0.085 |
| Oper1 | | 0.963 | 0.427 | 0.592 | 0.961 | 0.246 | 0.392 |
| macro average | | 0.506 | 0.534 | 0.341 | 0.503 | 0.515 | 0.239 |
| weighted average | | 0.923 | 0.437 | 0.570 | 0.921 | 0.270 | 0.379 |
| Oper2 & Oper3 | 3 | 0.028 | 0.628 | 0.053 | 0.027 | 0.796 | 0.053 |
| Oper1 | | 0.975 | 0.401 | 0.569 | 0.977 | 0.234 | 0.377 |
| macro average | | 0.502 | 0.515 | 0.311 | 0.502 | 0.515 | 0.215 |
| weighted average | | 0.950 | 0.407 | 0.555 | 0.952 | 0.249 | 0.369 |

Table 18 presents results for both methods averaged over classes and folds. The best precision was demonstrated by SVM, but its recall is as low as 0.274 resulting in F1-score of 0.387. SVM showed a higher recall of 0.427, and as its precision is not much lower than the NB precision, its F1-score of 0.564 is best. Table 19 give the results per class and fold, and it can be observed that NB detected Oper1 better that Oper2 and Oper3, showing a precision of 0.963, a recall of 427, and an F1-score of 0.592. SVM also distinguish Oper1 better than the other class with a precision of 0.965, a recall of 0.277, and an F1-score of 0.431.

### 5.3.5 Real

Real category includes two classes: the first class contains collocations of Real2 and Real3 put together in one class due to their similarity and a very small number of collocations: Real2 has 3 collocations in 2,942 sentences and Real3 has 1 collocation in 1,398 sentences, see Table 3. The second class includes Real1. Table 22 shows the dataset statistics for the classification, Table 23 includes overall averaged classification results for both classes, and Table 24 details the results for each fold, class, and algorithm.

Table 20. Dataset statistics for classification of Real collocations into two classes

| Class label | Fold | Train | | Test | |
|---|---|---|---|---|---|
| | | # collocations | # sentences | # collocations | # sentences |
| Real2 & Real3 | 1 | 2 | 4,005 | 2 | 335 |
| Real1 | | 41 | 21,294 | 20 | 6,946 |
| Real2 & Real3 | 2 | 3 | 4,184 | 1 | 156 |
| Real1 | | 40 | 19,501 | 21 | 8,739 |
| Real2 & Real3 | 3 | 3 | 4,161 | 1 | 179 |
| Real1 | | 41 | 15,685 | 20 | 12,555 |

Table 21. Overall averaged results for classifying Real collocations in two classes using Naïve Bayes (NB) and Support Vector Machine (SVM), best results in bold

| Model | Average weighted precision | Average weighted recall | Average weighted F1-score |
|---|---|---|---|
| NB | **0.849** | **0.218** | **0.234** |
| SVM | 0.847 | 0.187 | 0.186 |

Table 22. Results for classification of Real collocations into two classes using Naïve Bayes (NB) and Support Vector Machine (SVM), best results in bold

| Class label | Fold | Naïve Bayes | | | SVM | | |
|---|---|---|---|---|---|---|---|
| | | Precision | Recall | F1-score | Precision | Recall | F1-score |
| Real2 & Real3 | 1 | 0.419 | 0.798 | 0.550 | 0.397 | 0.747 | 0.518 |
| Real1 | | 0.757 | 0.363 | 0.490 | 0.702 | 0.345 | 0.463 |
| macro average | | 0.588 | 0.580 | 0.520 | 0.549 | 0.546 | 0.490 |
| weighted average | | 0.633 | 0.521 | 0.512 | 0.591 | 0.492 | 0.483 |
| Real2 & Real3 | 2 | 0.018 | 0.949 | 0.036 | 0.018 | 0.968 | 0.034 |
| Real1 | | 0.990 | 0.091 | 0.167 | 0.981 | 0.030 | 0.059 |
| macro average | | 0.504 | 0.520 | 0.101 | 0.500 | 0.500 | 0.047 |
| weighted average | | 0.973 | 0.106 | 0.165 | 0.964 | 0.047 | 0.058 |
| Real2 & Real3 | 3 | 0.014 | 0.955 | 0.027 | 0.014 | 1.000 | 0.028 |
| Real1 | | 0.953 | 0.013 | 0.026 | 1.000 | 0.008 | 0.016 |
| macro average | | 0.483 | 0.484 | 0.026 | 0.507 | 0.504 | 0.022 |
| weighted average | | 0.940 | 0.026 | 0.026 | 0.986 | 0.022 | 0.017 |

According to Table 21, the best results over classes and folds are given by NB: a precision of 0.849, a recall of 0.218, and an F1-score of 0.234. There is significant discrepancy between precision and recall producing a low F1-score. Although the precision of SVM is almost the same (0.847), its recall is much lower (0.187), so its F1-score of 0.186. Table 22 presents detailed results per class and per fold, here Real2 and Real3 class is best distinguished by both algorithms: NB produced a precision of 0.419, a recall of 0.798, and an F1-score of 0.550. The results for the same class demonstrated by SVM are 0.397, 0.747, 0.518 for precision, recall, and F1-score, respectively. It is notable here that both methods showed a recall higher than precision, unlike in Sections 5.3.1-5.3.4.

# 6 Discussion

First, we need to mention here, that we developed our classification methodology, chose BETO, a transformer trained on Spanish, and two machine learning methods for the experiments in order to showcase the utility of our dataset, whose compilation and description is the primary objective of this paper. Our results are given as an example of how the dataset can be applied on the one hand, and on the other hand, to create a baseline for further research and experimentation. However, the use of our dataset cannot be limited only to classification, it may serve for many other studies in linguistics and natural language processing.

We classified verb-noun phrases into collocations and free word combinations at Level 1 of our hierarchical scheme. Here the best result shown by BETO was an F1-score of 0.793 on collocation detection. At Level 2, on the task to classify collocations in ten lexical function types, the performance was lower than on Level 1, and best F1-score was 0.532 for the Func lexical function type.

At Level 3 we tested two algorithms: Naïve Bayes and Support Vector Machine because they are effective on many natural language processing tasks. However, the lexical function classification task showed to be hard for both methods, the best weighted average F1-score over all classes and folds of 0.747 was showed by SVM on the Func class, the highest result given by NB with the same measure was 0.667.

Level 3 classification into specific lexical functions was most difficult for NB and SVM. Among all NB results at this level, the best F1-score of 0.777 was shown for IncepOper1, and among all SVM results, the best F1-score of 0.782 was for detecting Func0 by SVM. In general, in most cases, precision was significantly higher than recall resulting in low F1-score values.

Concluding this work, we suggest to continue research on lexical function classification task with our dataset applying other machine learning techniques and language models. This will contribute to improving semantic analysis in natural language systems and applications for machine translation, text generation, language understanding, among many other objectives.

# 7 Conclusion

This paper presented a new dataset of 957 frequent Spanish verb-noun phrases, 737 of which are collocations and 220 phrases are free word combinations. For each phrase, all sentences with its occurrence were extracted from *Excelsior*, *La Razón*, and *Público* newspapers by parsing dependency trees and from. All collocations in the dataset were annotated with lexical functions of the Meaning-Text Theory (Mel'čuk 2015).

To showcase the use of our dataset, we presented a hierarchical classification task, where each verb-noun phrase was classified at three levels: at the first level, all phrases were classified in two classes: collocations and free word combinations, further at the other levels only collocations were classified according to lexical functions on a coarse-grained basis at the second level and consequently on a fine-grained basis at the third level of the hierarchical classification. As features for the classification, the words surrounding the verb-noun phrases in sentences was used, therefore, this task required a good understanding of context and relationships between words. We provide baselines and data splits for each classification level.

This work aims to contribute to the ongoing research efforts in the field of natural language processing and support the development of models with improved performance in recognizing collocations and their lexical functions.

**Acknowledgments**

The work was done with partial support from the Mexican Government through the grant A1S-47854 of CONACYT, Mexico, grants 20232138, 20231567, and 20232080 of the Secretaría de Investigación y Posgrado of the Instituto Politécnico Nacional, Mexico. The authors thank the CONACYT for the computing resources brought to them through the Plataforma de Aprendizaje Profundo para Tecnologías del Lenguaje of the Laboratorio de Supercómputo of the INAOE, Mexico, and acknowledge the support of Microsoft through the Microsoft Latin America PhD Award.